\pdfoutput=1

\documentclass[runningheads,a4paper]{llncs}

\usepackage{amssymb}
\usepackage{graphicx}
\usepackage{tabularx}
\usepackage{multirow}

\def\etal{\emph{et al. }}
\newcolumntype{Y}{>{\centering\arraybackslash}X}

\usepackage{url}
\urldef{\mailsa}\path|{alfred.hofmann, ursula.barth, ingrid.haas, frank.holzwarth,|
\urldef{\mailsb}\path|anna.kramer, leonie.kunz, christine.reiss, nicole.sator,|
\urldef{\mailsc}\path|erika.siebert-cole, peter.strasser, lncs}@springer.com|    
\newcommand{\keywords}[1]{\par\addvspace\baselineskip
\noindent\keywordname\enspace\ignorespaces#1}

\begin{document}

\mainmatter  

\title{Transitioning between Convolutional and Fully Connected Layers in Neural Networks}

\titlerunning{Transition Module}

%
%
\author{Shazia Akbar \inst{1}, Mohammad Peikari \inst{1}, Sherine Salama \inst{2}, Sharon Nofech-Mozes \inst{2}, Anne Martel \inst{1}}

%
\authorrunning{Akbar \etal}
\institute{Sunnybrook Research Institute, University of Toronto, Toronto \\ \email{\{sakbar,mpeikari,amartel\}@sri.utoronto.ca} \and Department of Pathology, Sunnybrook Health Sciences Centre, Toronto \\ \email{\{sherine.salama,sharon.nofech-mozes\}@sunnybrook.ca}}


%
%

\toctitle{Sparse Coding Layer}
\tocauthor{Akbar \etal}
\maketitle

\begin{abstract}
Digital pathology has advanced substantially over the last decade however tumor localization continues to be a challenging problem due to highly complex patterns and textures in the underlying tissue bed. The use of convolutional neural networks (CNNs) to analyze such complex images has been well adopted in digital pathology. However in recent years, the architecture of CNNs have altered with the introduction of inception modules which have shown great promise for classification tasks. In this paper, we propose a modified ``transition'' module which learns global average pooling layers from filters of varying sizes to encourage class-specific filters at multiple spatial resolutions. We demonstrate the performance of the transition module in AlexNet and ZFNet, for classifying breast tumors in two independent datasets of scanned histology sections, of which the transition module was superior.

\keywords{Convolutional neural networks, histology, transition, inception, breast tumor}
\end{abstract}

\section{Introduction}
There are growing demands for tumor identification in pathology for time consuming tasks such as measuring tumor burden, grading tissue samples, determining cell cellularity and many others. Recognizing tumor in histology images continues to be a challenging problem due to complex textural patterns and appearances in the tissue bed. With the addition of tumor, subtle changes which occur in the underlying morphology are difficult to distinguish from healthy structures and require expertise from a trained pathologist to interpret. An accurate automated solution for recognizing tumor in vastly heterogeneous pathology datasets would be of great benefit, enabling high-throughput experimentation, greater standardization and easing the burden of manual assessment of digital slides.

Deep convolutional neural networks (CNNs) are now a widely adopted architecture in machine learning. Indeed, CNNs have been adopted for tumor classification in applications such as analysis of whole slide images (WSI) of breast tissue using AlexNet \cite{Spanhol2016} and voxel-level analysis for segmenting tumor in CT scans \cite{Vivanti2015}. Such applications of CNNs continue to grow and the traditional architecture of a CNN has also evolved since its origination in 1998 \cite{LeCun1998}. A basic CNN architecture encompasses a combination of convolution and pooling operations. As we traverse deeper in the network, the network size decreases resulting in a series of outputs, whether that be classification scores or regression outcomes. In lower layers of a typical CNN, fully-connected (FC) layers are required to learn non-linear combinations of learned features. However the transition between a series of two-dimensional convolutional layers to a one-dimensional FC layer is abrupt, making the network susceptible to overfitting \cite{Lin2014}. In this paper we propose a method for transitioning between convolutional layers and FC layers by introducing a framework which encourages generalization. Different from other regularizers \cite{Srivastava2014,Krizhevsky2012}, our method congregates high-dimensional data from features maps produced in convolutional layers in an efficient manner before flattening is performed.

To ease the dimensionality reduction between convolutional and FC layers we propose a transition module, inspired by the inception module \cite{Szegedy2015}. Our method encompasses convolution layers of varying filter sizes, capturing learned feature properties at multiple scales, before collapsing them to a series of average pooling layers. We show that this configuration gives considerable performance gains for CNNs in a tumor classification problem in scanned images of breast cancer tissue. We also evaluate the performance of the transition module compared to other commonly used regularizers (section \ref{sec:r_regularizers}).

\begin{figure}[t]
	\centering
	\begin{tabular}{ccc}
		{\includegraphics[width=0.32\textwidth]{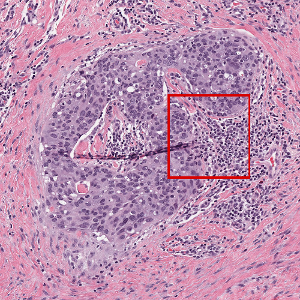}} &
		{\includegraphics[width=0.32\textwidth]{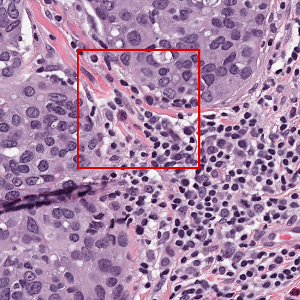}} &
		{\includegraphics[width=0.32\textwidth]{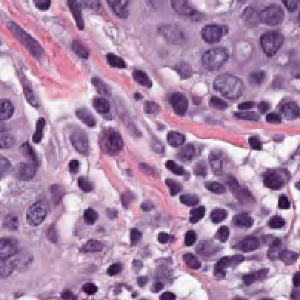}} \\
	\end{tabular}
	\caption{Digital slide shown at multiple resolutions. Regions-of-interest outlined in red are shown at greater resolutions from left to right.}
	\label{fig:multi}
\end{figure}

\section{Related Work}

In the histopathology literature, CNN architectures tend to follow a linear trend with a series of layers sequentially arranged from an input layer to a softmax output layer \cite{Spanhol2016,Litjens2016}. Recently, however, there have been adaptations to this structure to encourage multiscale information at various stages in the CNN design. For example Xu \etal \cite{Xu2016} proposed a multichannel side supervision CNN which merges edges and contours of glands at each convolution layer for gland segmentation. In cell nuclei classficiation, Buyssens \etal \cite{Buyssens2013} learn multiple CNNs in parallel with input images at  various resolutions before aggregating classification scores. These methods have shown to be particularly advantageous in histology images as it mimics pathologists' interpretation of digital slides when viewed at multiple objectives (Fig. \ref{fig:multi}).

However capturing multiscale information in a single layer has been a recent advancement after the introduction of the inception modules (Section \ref{sec:inceptiondesc}). Since then, there have been some adaptations however these have been limited to convolution layers in a CNN. Liao and Carneiro \cite{Liao2015} proposed a module which combines multiple convolution layers via a max-out operation as opposed to concatenation. Jin \etal \cite{Jin2016} designed a CNN network in which independent FC layers are learned from the outputs of inception layers created at various levels of the network structure. In this paper, we focus on improving the network structure when changes in dimensionality occur between convolution and FC layers. Such changes occur when the network has already undergone substantial reduction and approaches the final output layer therefore generalization is key for optimal class separation.

\section{Method}

\subsection{Inception Module}
\label{sec:inceptiondesc}

\begin{figure}
	\centering
	\includegraphics[width=\textwidth]{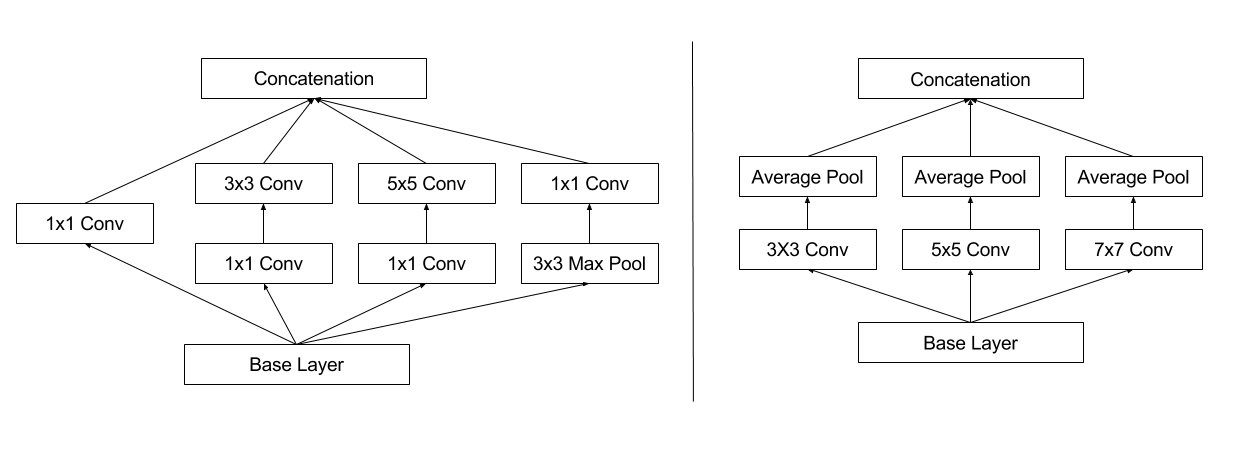}
	\caption{Original inception module (left) and the proposed transition module (right).}
	\label{fig:compare}
\end{figure}

Inception modules, originally proposed by Szegedy \etal \cite{Szegedy2015}, are a method of gradually increasing feature map dimensionality thus increasing the depth of a CNN without adding extreme computational costs. In particular, the inception module enables filters to be learned at multiple scales simultaneously in a single layer, also known as a sub-network. Since its origination, there have been multiple inception networks incorporating various types of inception modules \cite{Szegedy2016,Szegedy2016b}, including GoogleNet. The base representation of the original inception module is shown in Fig. \ref{fig:compare} (left). 

Each inception module is a parallel series of convolutional layers restricted to filter sizes $1$x$1$, $3$x$3$ and $5$x$5$. By encompassing various convolution sub-layers in a single deep layer, features can be explored at multiple scales simultaneously. During training, the combination of filter sizes which result in optimal performance are weighted accordingly.
However on its own, this configuration results in a very large network with increased complexity. Therefore for practicality purposes, the inception module also encompasses $1$x$1$ convolutions which act as dimensionality reduction mechanisms. The Inception \emph{network} is defined as a stack of inception modules with occasional max-pooling operations. The original implementation of the Inception network encompasses nine inception modules \cite{Szegedy2015}.

\subsection{Transition Module}

In this paper, we propose a modified inception module, called the ``transition'' module, explicitly designed for the final stages of a CNN, in which learned features are mapped to FC layers. Whilst the inception module successfully captures multiscale from input data, the bridge between learned feature maps and classification scores is still treated as a black box. To ease this transition process we propose a method for enabling 2D feature maps to be downscaled substantially before tuning FC layers.
In the transition module, instead of concatenating outcomes from each filter size, as in \cite{Szegedy2015}, independent global average pooling layers are configured after learning convolution layers which enable feature maps to be compressed via an average operation.

Originally proposed by Lin \etal \cite{Lin2014}, global average pooling layers were introduced as a method of enforcing correspondences between categories of the classification task (i.e. the softmax output layer) and filter maps. As the name suggests, in a global averaging pooling layer a single output is retrieved from each filter map in the preceeding convolutional layer by averaging it. For example, if given an input of $256$ $3$x$3$ feature maps, a global average pool layer would form an output of size $256$. 
In the transition module, we use global average pooling to sum out spatial information at multiple scales before collapsing each averaged filter to independent $1$D output layers. This approach has the advantage of introducing generalizability and encourages a more gradual decrease in network size earlier in the network structure. As such, subsequent FC layers in the network are also smaller in size, making the task of delineating classification categories much easier. Furthermore there are no additional parameters to tune. 

The structure of the transition module is shown in Fig. \ref{fig:compare} (right). Convolution layers were batch normalized as proposed in Inception-v2 for further regularization.

\section{Experiment}


We evaluated the performance of the proposed transition module using a dataset of $1229$ image patches extracted and labelled from breast WSIs scanned at x$20$ magnification by a Scanscope XT (Aperio technologies, Leica Biosystems) scanner. Each RGB patch of size $512$x$512$ was hand selected from $31$ WSIs, each one from a single patient, by a trained pathologist. Biopsies were extracted from patients with invasive breast cancer and subsequently received neo-adjuvant therapy; post neoadjuvant tissue sections revealed invasive and/or ductal carcinoma \emph{in situ}. 
$5$-fold cross validation was used to evaluate the performance of this dataset. Each image patch was confirmed to contain either tumor or healthy tissue by an expert pathologist. ``Healthy'' refers to patches which are absent of cancer cells but may contain healthy epithelial cells amongst other tissue structures such as stroma, fat etc. Results are reported over $100$ epochs. 

We also validated our method on a public dataset, BreaKHis \cite{Spanhol2016b} (section \ref{sec:breakhist}) which contains scanned images of benign (adenosis, fibroadenoma, phyllodes tumor, tubular adenoma) and malignant (ductal carcinoma, lobular carcinoma, mucinous carcinoma, papillary carcinoma) breast tumors at x$40$ objective. Images were resampled into patches of dimensions $228$x$228$, suitable for a CNN, resulting in $11,800$ image patches in total. BreaKHis was validated using $2$-fold cross validation and across $30$ epochs. 


In section \ref{sec:r_architecture}, results are reported for three different CNN architectures (AlexNet, ZFNet, Inception-v3), of which transition modules were introduced in AlexNet and ZFNet. All CNNs were trained from scratch with no pretraining. Transition modules in both implementations encompassed $3$x$3$, $5$x$5$ and $7$x$7$ convolutions, thus producing three average pooling layers. Each convolutional layer has a stride of $2$, and $1024$ and $2048$ filter units for AlexNet and VFNet, respectively. Note, the number of filter units were adapted according to the size of the first FC layer proceeding the transition module.

CNNs were implemented using Lasagne $0.2$ \cite{Lasagne}. A softmax function was used to obtain classification predictions and convolutional layers encompassed Re{LU} activations. $10$ training instances were used in each batch in both datasets. We used Nestorov Momentum \cite{Sutskever2013} to perform updates with a learning rate of $1e^{-5}$.

\section{Results}


\subsection{Experiment 1: Comparison with Regularizers}
\label{sec:r_regularizers}

Our first experiment evaluated the performance of the transition model when compared to other commonly used regularizers including Dropout \cite{Srivastava2014} and cross-channel local response normalization \cite{Krizhevsky2012}. We evaluated the performance of each regularizer in AlexNet and report results for a) a single transition module added before the first FC layer, b) two Dropout layers, one added after each FC layer with $p=0.5$, and lastly c) normalization layers added after each max-pooling operation, similar to how it was utilized in \cite{Krizhevsky2012}. 

The transition module achieved an overall accuracy rate of $91.5\%$ which when compared to Dropout ($86.8\%$) and local normalization ($88.5\%$) showed considerable improvement, suggesting the transition module makes an effective regularizer compared to existing methods. When local response normalization was used in combination with the transition module in ZFNet (below), we achieved a slightly higher test accuracy of $91.9\%$.

\subsection{Experiment 2: Comparing Architectures}
\label{sec:r_architecture}

\begin{figure}[t]
	\centering
	\includegraphics[width=1.0\textwidth]{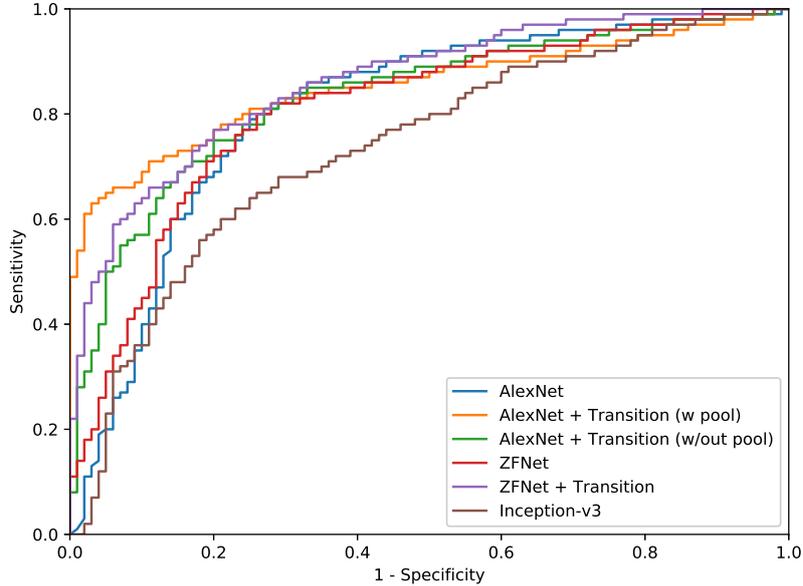}
	\caption{ROC curves for AlexNet \cite{Krizhevsky2012} and ZFNet \cite{Zeiler2013}, with and without the proposed transition module, and Inception-v3 \cite{Szegedy2016}. ROC curves are also shown for the transition module with and without average pooling.}
	\label{fig:roc}
\end{figure}

Next we evaluated the performance of the transition module in two different CNN architectures: AlexNet and ZFNet. We also report the performance of Inception-v3 which already has built-in regularizers in the form of $1$x$1$ convolutions \cite{Szegedy2016}, for comparative purposes. ROC curves are shown in Fig. \ref{fig:roc}.


Both AlexNet and ZFNet benefited from the addition of a single transition module, improving test accuracy rates by an average of $4.3\%$, particularly at lower false positive rates. Smaller CNN architectures proved to be better for tumor classification in this case as overfitting was avoided, as shown by the comparison with Inception-v3. Surprisingly, the use of dimensionality reduction earlier in the architectural design does not prove to be effective for increasing classification accuracy.
We also found that the incorporation of global average pooling in the transition module improved results slightly and resulted in $3.1\%$ improvement in overall test accuracy.

\subsection{Experiment 3: BreaKHis}
\label{sec:breakhist}

\begin{figure}[t]
	\centering
	\includegraphics[width=0.9\textwidth]{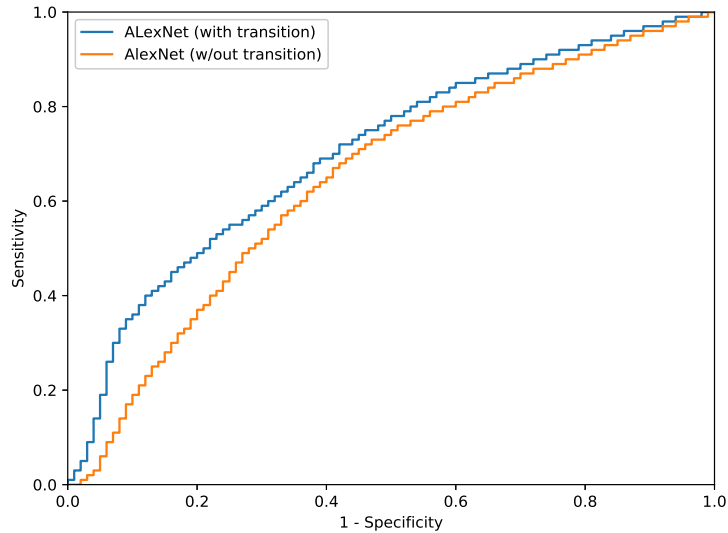}
	\caption{ROC curves for BreaKHis \cite{Spanhol2016b} dataset with and without the proposed transition module.}
	\label{fig:breakhist_roc}
\end{figure}

We used the same AlexNet architecture used above to also validate BreaKHis. ROC curves are shown in Fig. \ref{fig:breakhist_roc}. There was a noticeable improvement (AUC+=$0.06$) when the transition module was incorporated, suggesting that even when considerably more training data is available a smoother network reduction can be beneficial.

The transition module achieved an overall test accuracy of $82.7\%$ which is comparable to $81.6\%$ achieved with SVM in \cite{Spanhol2016b}, however these results are not directly comparable and should be interpreted with caution.

\section{Conclusion}

In this paper we propose a novel regularization technique for CNNs called the transition module, which captures filters at multiple scales and then collapses them via average pooling in order to ease network size reduction from convolutional layers to FC layers. We showed that in two CNNs (AlexNet, ZFNet) this design proved to be beneficial for distinguishing tumor from healthy tissue in digital slides. We also showed an improvement in a larger publically available dataset, BreaKHis. 

\section*{Acknowledgements}
This work has been supported by grants from the Canadian Breast Cancer Foundation, Canadian Cancer Society (grant 703006) and the National Cancer Institute of the National Institutes of Health (grant number U24CA199374-01). 

\bibliographystyle{splncs03}
\bibliography{references}

\end{document}